\documentclass[runningheads]{llncs}
\usepackage[T1]{fontenc}
\usepackage{graphicx}
\usepackage{caption}
\usepackage{subcaption}
\usepackage{multicol}
\usepackage{breakcites} 
\usepackage{booktabs}
\usepackage{multirow, makecell}

\usepackage[english]{babel}
\usepackage{xcolor,colortbl}
\usepackage{blindtext}

\usepackage{hyperref}
\hypersetup{
    colorlinks=true,
    linkcolor=blue,
    filecolor=magenta,      
    urlcolor=magenta,
    pdftitle={AnnoPage Dataset: Dataset of Non-Textual Elements in Documents with Fine-Grained Categorization},
}

\begin{document}
\title{AnnoPage Dataset: Dataset of Non-Textual Elements in Documents with Fine-Grained Categorization}
\titlerunning{AnnoPage Dataset}
\author{ 
Martin Kišš\inst{1} \orcidID{0000-0001-6853-0508} \and
Michal Hradiš\inst{1} \orcidID{0000-0002-6364-129X} \and
Martina Dvořáková\inst{2} \and
Václav Jiroušek\inst{3} \and
Filip Kersch\inst{4}
}
\authorrunning{M. Kišš et al.}
\institute{
Faculty of Information Technology, Brno University of Technology, \\
Brno, Czech Republic\\
\email{\{ikiss,hradis\}@fit.vutbr.cz} \vspace{0.2cm}
\and
Moravian Library, \\
Brno, Czech Republic\\
\email{Martina.Dvorakova@mzk.cz} \vspace{0.2cm}
\and
National Library of the Czech Republic, \\
Prague, Czech Republic\\
\email{Vaclav.Jirousek@nkp.cz} \vspace{0.2cm}
\and
Library of the Czech Academy of Sciences, \\
Prague, Czech Republic\\
\email{kersch@lib.cas.cz}
}
\maketitle              % typeset the header of the contribution
\begin{abstract}

We introduce the AnnoPage Dataset, a novel collection of 7,550 pages from historical documents, primarily in Czech and German, spanning from 1485 to the present, focusing on the late 19th and early 20th centuries. 
The dataset is designed to support research in document layout analysis and object detection. 
Each page is annotated with axis-aligned bounding boxes (AABB) representing elements of 25 categories of non-textual elements, such as images, maps, decorative elements, or charts, following the Czech Methodology of image document processing. 
The annotations were created by expert librarians to ensure accuracy and consistency.
The dataset also incorporates pages from multiple, mainly historical, document datasets to enhance variability and maintain continuity. 
The dataset is divided into development and test subsets, with the test set carefully selected to maintain the category distribution.
We provide baseline results using YOLO and DETR object detectors, offering a reference point for future research. 
The AnnoPage Dataset is publicly available on Zenodo\footnote{\url{https://doi.org/10.5281/zenodo.12788419}}, along with ground-truth annotations in YOLO format.

\keywords{Dataset \and Non-Textual Elements \and Graphical Elements \and Documents.}
\end{abstract}
\section{Introduction}
In recent years, the digitization of documents in various institutions has made the content of documents accessible to the general public.
However, while the textual content of documents is usually easily recognizable using today's text recognition technologies (OCR), other document content is often not processed in libraries and archives, even though it may contain important information.
Such content includes images, photographs, charts, various diagrams and drawings, musical notation, etc.
The first and key step in processing such elements is the ability to detect them on the page and determine their category.
Subsequently, these elements can be processed according to their category; for instance, images can be indexed for search, musical notations interpreted using optical music recognition (OMR), and mathematical expressions analyzed through math recognition systems. 
Such category-specific processing greatly expands the possibilities for users interacting with digitised documents, benefiting both general audiences and specialists conducting targeted research.

For document layout analysis and object detection in documents, many datasets exist~\cite{pfitzmann_doclaynet_2022,zhong_publaynet_2019,antonacopoulos_realistic_2009,tschirschwitz_dataset_2022,jaume2019funsd,diem_cbad_2019,clausner_enp_2015,li_docbank_2020,smock_pubtables-1m_2022,cheng_m6doc_2023}.
However, these datasets contain only several categories of elements~\cite{zhong_publaynet_2019,monnier_docextractor_2020,clausner_enp_2015}, they focus mainly on identification of different parts of textual content, such as captions, titles, headers, footers, etc.~\cite{pfitzmann_doclaynet_2022,ma_hrdoc_2023,li_docbank_2020}, they are synthetically generated~\cite{monnier_docextractor_2020}, or they focus only on modern documents~\cite{antonacopoulos_realistic_2009,zhong_publaynet_2019,pfitzmann_doclaynet_2022,ma_hrdoc_2023,li_docbank_2020}.
Some datasets focus only on a single category, such as text lines~\cite{diem_cbad_2019,kodym_page_2021} or tables~\cite{shahab_open_2010,smock_pubtables-1m_2022,li_tablebank_2020,gao_icdar_2019}.
Also, historical documents, preserved by numerous institutions, also contain unique visual elements typically absent from modern documents, such as friezes, vignettes, signets, and other decorative components.

\begin{figure}[t]
    \centering
    \includegraphics[width=0.32\linewidth]{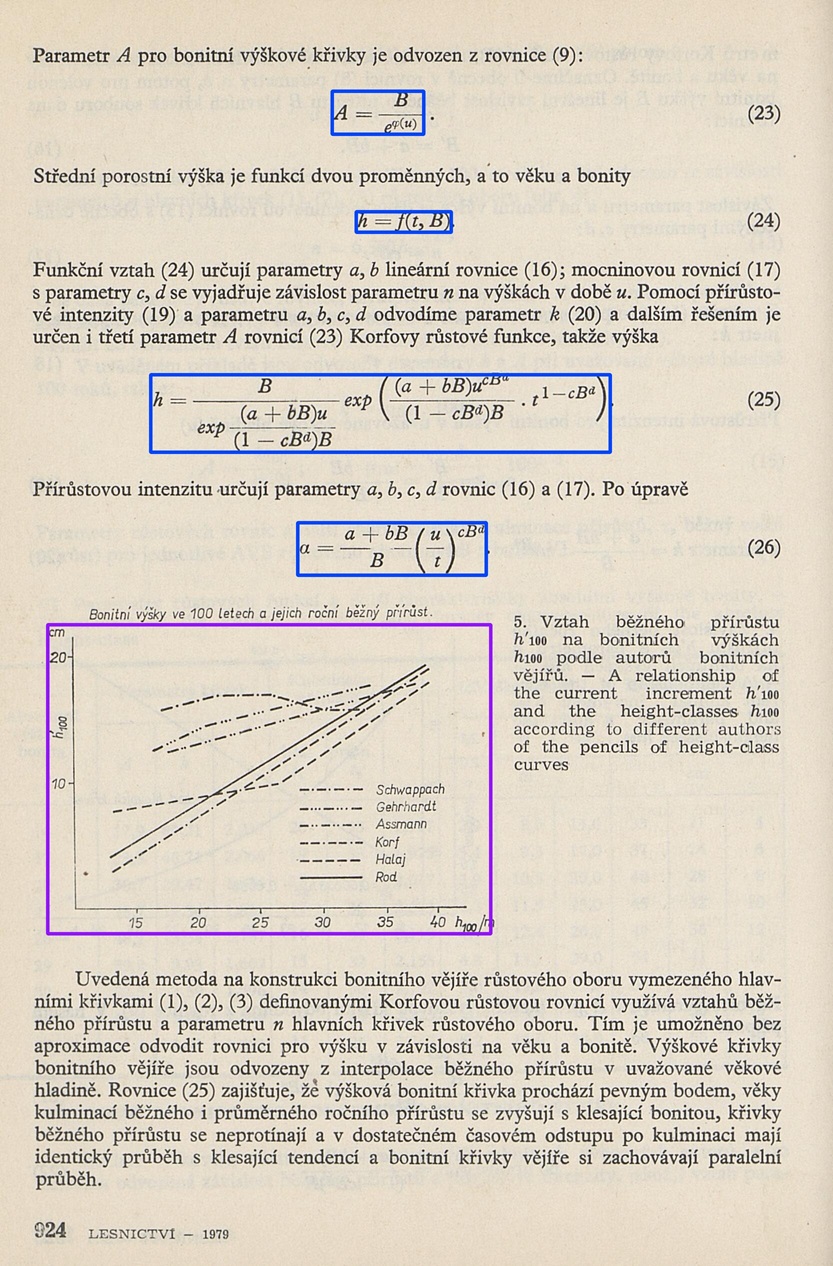}
    \hfill
    \includegraphics[width=0.32\linewidth]{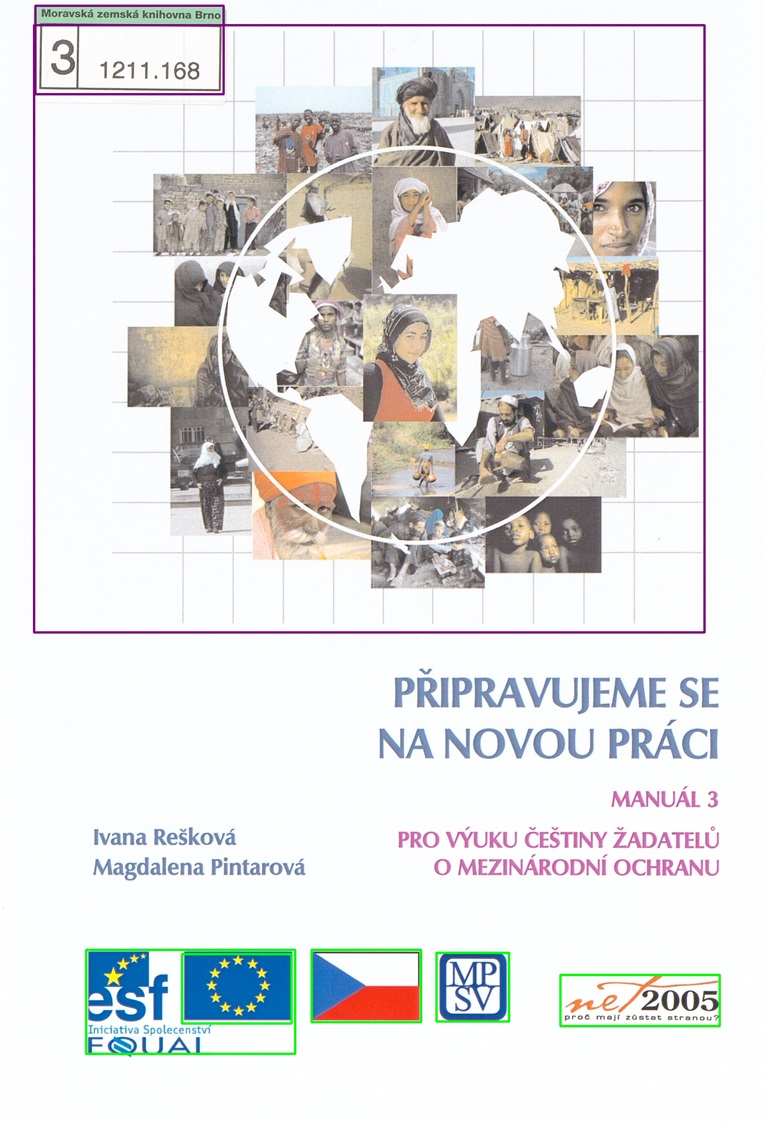}
    \hfill
    \includegraphics[width=0.32\linewidth]{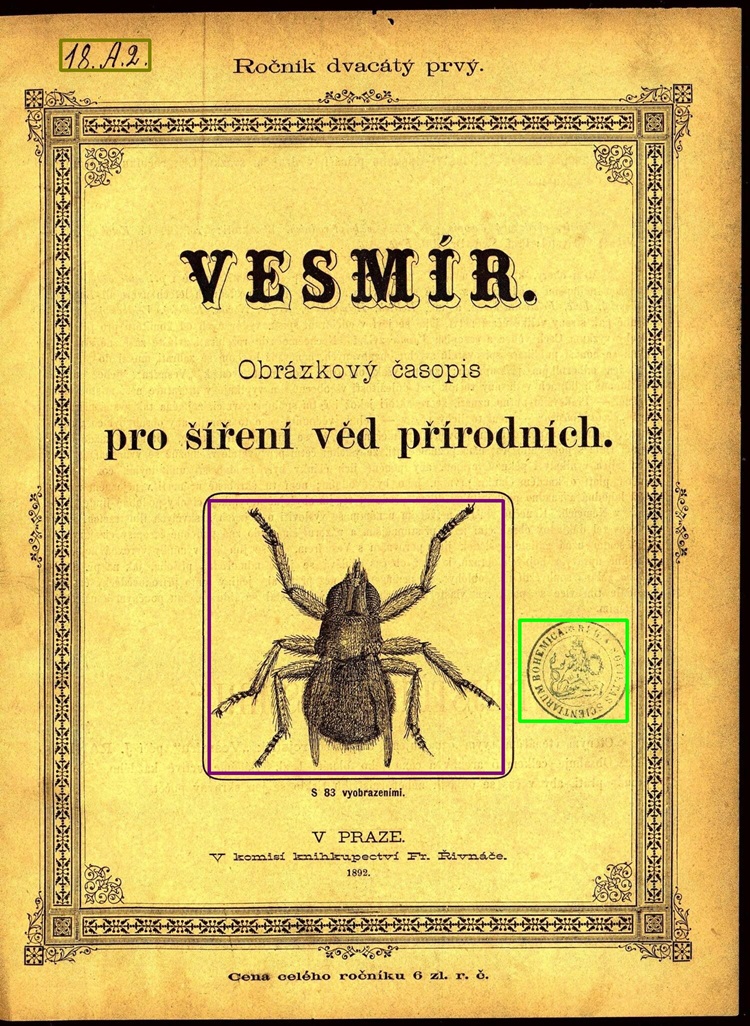}
    \\
    \includegraphics[width=0.32\linewidth]{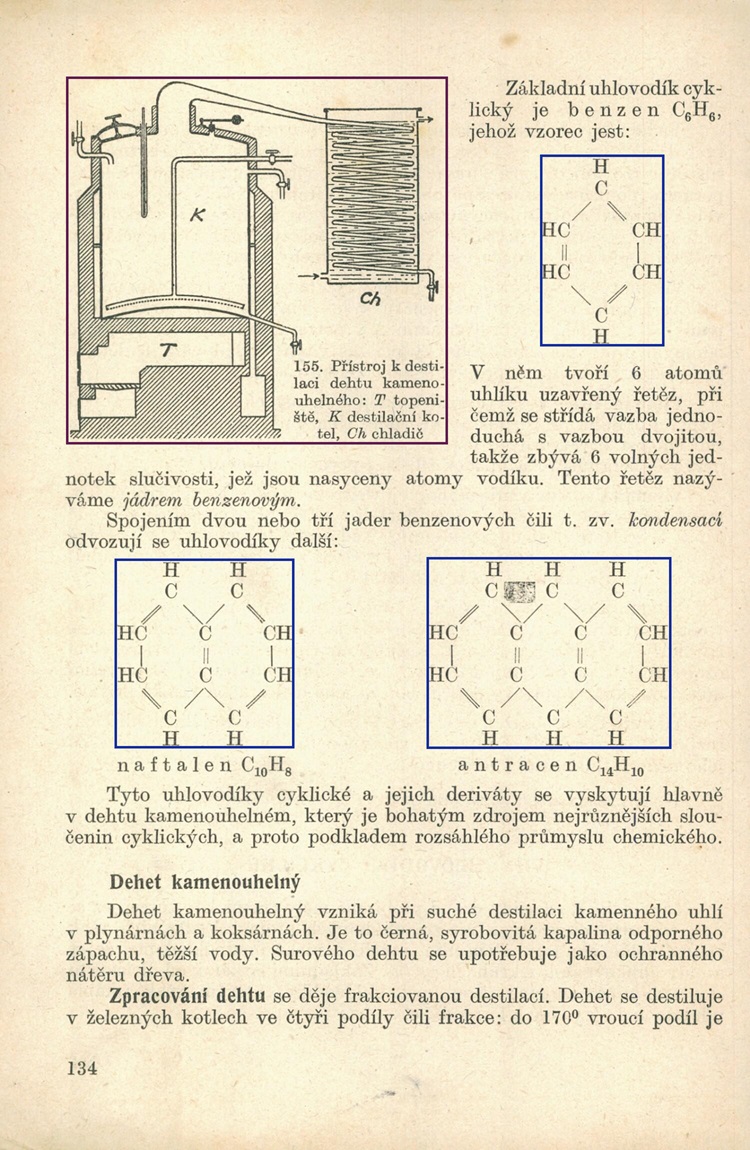}
    \hfill
    \includegraphics[width=0.32\linewidth]{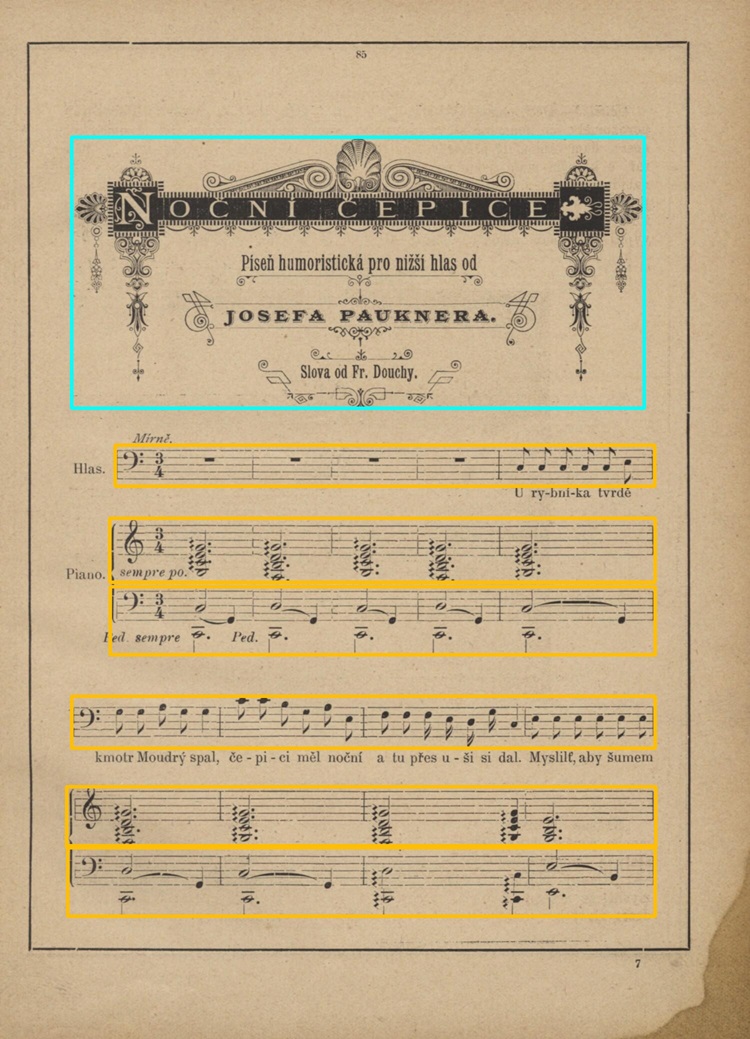}
    \hfill
    \includegraphics[width=0.32\linewidth]{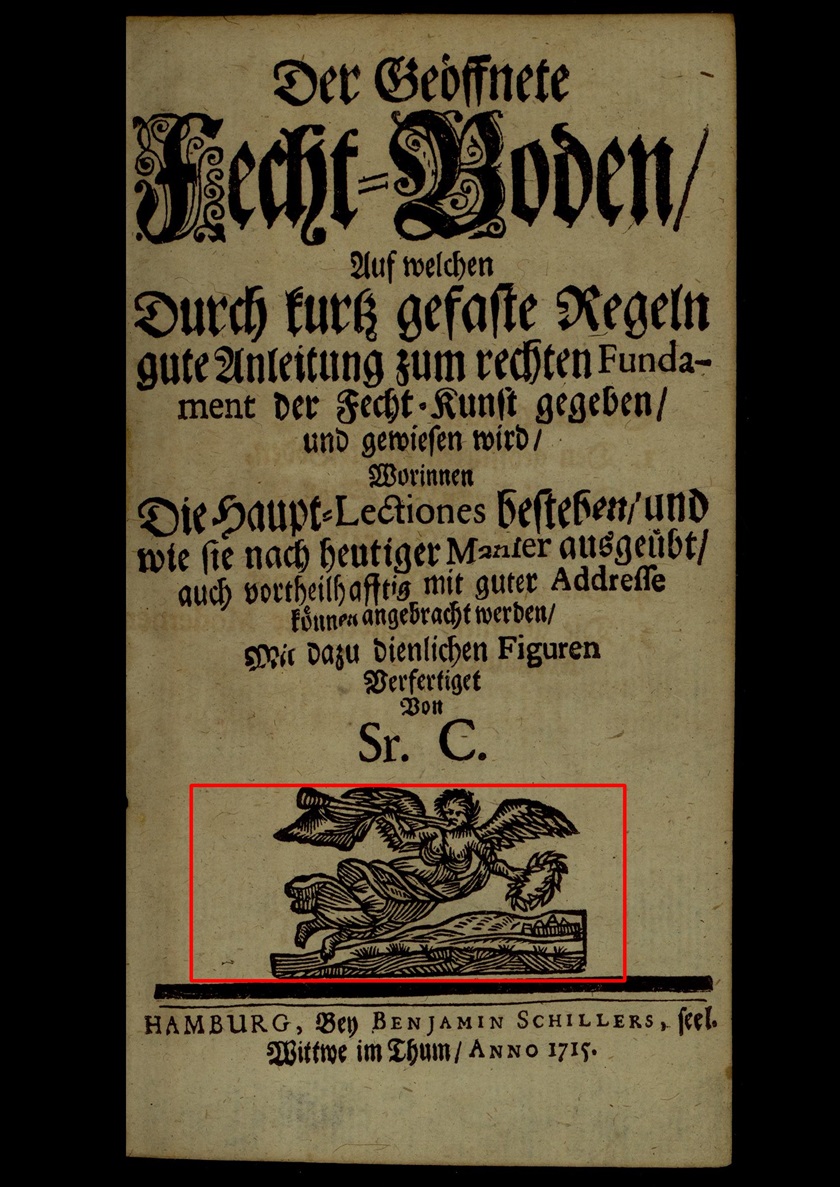}
    \caption{Example pages from the AnnoPage Dataset with rendered ground-truth in colored bounding-boxes.}
    \label{fig:examples}
\end{figure}

In this paper, we present the novel AnnoPage Dataset, which contains $7\,550$ pages from various, mainly historical documents.
The pages were written mostly in Czech and German at the end of the 19th and during the 20th century (see Figure~\ref{fig:examples} for example pages).
These pages were carefully annotated by expert librarian annotators who used axis-aligned bounding boxes to identify elements of 25 categories of non-textual elements according to the Czech Methodology of image document processing~\cite{jirousek_metodika_2024}.

The contributions of this paper are as follows:
\begin{enumerate}
    \item We introduce the AnnoPage Dataset, its creation process, limitations, and detailed information about the source documents and the annotations.
    \item We present baseline results based on YOLO and DETR object detectors.
\end{enumerate}

\section{Related work}
\label{sec:datasets}

In general, document layout analysis is a task for which a large number of diverse datasets exist~\cite{kodym_page_2021,diem_cbad_2019,soboroff2022complex,jaume2019funsd,pfitzmann_doclaynet_2022,zhong_publaynet_2019,cheng_m6doc_2023}.
There are also datasets that focus on detecting various non-textual elements such as images, tables, or mathematical formulas~\cite{antonacopoulos_realistic_2009,clausner_enp_2015,boillet_horae_2019,clausner_icdar2019_2019,monnier_docextractor_2020,buttner_cordeep_2022,tschirschwitz_dataset_2022,zhong_publaynet_2019,pfitzmann_doclaynet_2022,ma_hrdoc_2023,li_docbank_2020}.
In the terms of annotations, the ground-truth typically contains elements annotated either as bounding boxes~\cite{boillet_horae_2019,buttner_cordeep_2022,tschirschwitz_dataset_2022,zhong_publaynet_2019,pfitzmann_doclaynet_2022,li_tablebank_2020,ma_hrdoc_2023} or bounding polygons~\cite{antonacopoulos_realistic_2009,clausner_enp_2015,monnier_docextractor_2020,clausner_icdar2019_2019}.
Some datasets contain pixel-level annotations, such as HBA 1.0 dataset~\cite{mehri_hba_2017}, DIVA-HisDB~\cite{simistira_diva-hisdb_2016}, or U-DIADS-Bib~\cite{zottin_u-diads-bib_2024}.

One of the existing datasets designed for document layout analysis that has a different type of annotation is the , which contains pixel-level annotations.

From the existing datasets, the closely related ones to the AnnoPage Dataset are PRImA Layout Analysis Dataset~\cite{antonacopoulos_realistic_2009}, HORAE~\cite{boillet_horae_2019}, S-VED~\cite{buttner_cordeep_2022}, DocLayNet~\cite{pfitzmann_doclaynet_2022}, and TexBiG~\cite{tschirschwitz_dataset_2022}.
The PRImA Layout Analysis Dataset contains pages from contemporary documents annotated with tight bounding polygons for texts, images, line drawings, graphics, tables, charts, separators, maths, noise, and frames.
Similarly, the DocLayNet dataset consists of contemporary documents, but the annotations are more focused on various types of texts, such as captions, titles, headers, footers, etc.
In terms of non-textual elements, the dataset contains bounding box annotations for images, mathematical formulas, and tables.

The HORAE dataset contains pages from historical documents and books with annotated bounding boxes for texts, images, borders, initials, music notations, and ornaments.
Similarly, the S-VED dataset is a collection of historical pages with annotations of illustrations, decorations, printer's marks, and initials in the form of bounding boxes.
Probably the most closely related dataset to the AnnoPage Dataset is the TexBiG dataset.
It contains pages from six historical books from the 19th and early 20th century written in German.
The ground-truth contains annotations for advertisements, decorations, equations, frames, images, logos, tables, and several categories for text.

\subsection{Approaches to object detection in documents}

The approaches to object detection in documents depend mainly on the type of the annotations.
For bounding polygon-based and pixel-level annotations, the approaches often utilize various segmentation neural networks~\cite{monnier_docextractor_2020,buttner_cordeep_2022,boillet_horae_2019}, such as U-Net and its modifications.
In the case of bounding box annotations, the approaches are typically based on object detectors such as R-CNN~\cite{auer_icdar_2023,tom_icdar_2023,pfitzmann_doclaynet_2022,li_docbank_2020,ma_hrdoc_2023,zhong_publaynet_2019} and its variants (Mask R-CNN, Faster R-CNN, Cascade R-CNN), YOLO~\cite{zhao_doclayout-yolo_2024,auer_icdar_2023,buttner_cordeep_2022}, and DETR~\cite{carion_end--end_2020,tom_icdar_2023,auer_icdar_2023} and its modifications (DeformableDETR~\cite{zhu_deformable_2021}, DINO~\cite{zhang_dino_2022}, and MaskDINO~\cite{li_mask_2023}).

\section{AnnoPage Dataset}
\label{sec:annopage}
The AnnoPage Dataset aims to create a collection of pages annotated with non-textual elements with fine-grained categorization.
The dataset was annotated mainly by librarians who followed the Czech Methodology of image document processing~\cite{jirousek_metodika_2024}.
As a result, the dataset contains $7\,550$ pages of mostly historical documents annotated with $27\,904$ elements where each element is represented as an axis-aligned bounding box {AABB}.
In the following subsections, we describe the entire process of creating this dataset and the evaluation protocol associated with the dataset.

\subsection{Data acquisition}
\begin{figure}[t]
    \centering
    \includegraphics{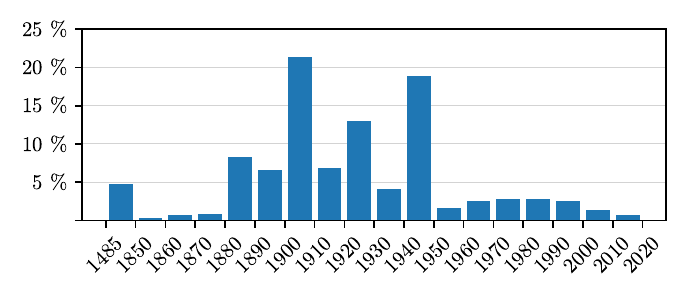}
    \caption{Distribution of publication year of pages in the dataset.}
    \label{fig:dates}
\end{figure}

We have collected $7\,550$ pages from various documents.
Most of the pages ($5\,690$) come from publicly accessible documents from the Czech Digital Library\footnote{\url{https://www.digitalniknihovna.cz/}}.
These pages are written mostly in Czech and German, but there are also a few pages containing other languages, such as French, and English.
In terms of dating, the pages are from as early as 1485 to the present, with the majority from the late 19th and the first half of the 20th centuries.
The distribution of the pages over the years is shown in Figure~\ref{fig:dates}.

The rest of the pages come from existing datasets, most of them containing historical pages, namely IlluHisDoc~\cite{monnier_docextractor_2020} (237 pages), PRImA Layout Analysis Dataset~\cite{antonacopoulos_realistic_2009} (170 pages), PRImA RDCL2019~\cite{clausner_icdar2019_2019} (40 pages), PRImA Europeana Newspapers~\cite{clausner_enp_2015} (361 pages), ICDAR2019 cBAD dataset~\cite{diem_cbad_2019} (382 pages), and TexBiG~\cite{tschirschwitz_dataset_2022} (670 pages).
The purpose of these extra pages is to increase the variability of the final dataset and also to maintain continuity across datasets.
Although these datasets contain the original ground-truth, we have created new annotations for these pages following the Czech Methodology of image document processing~\cite{jirousek_metodika_2024}.
To distinguish the Czech data from the pages from the extra datasets and the full dataset, we refer to the Czech data as the \emph{Czech subset} in the rest of the paper.

\subsection{Element categories}
\label{sec:categories}
\begin{figure}[t]
    \centering
    \begin{subfigure}{0.465\linewidth}
        \centering
        \includegraphics[width=0.48\linewidth]{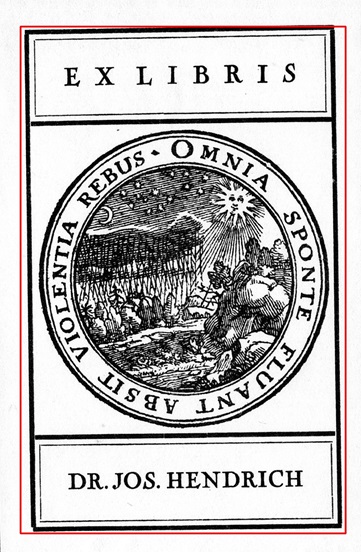}
        \includegraphics[width=0.48\linewidth]{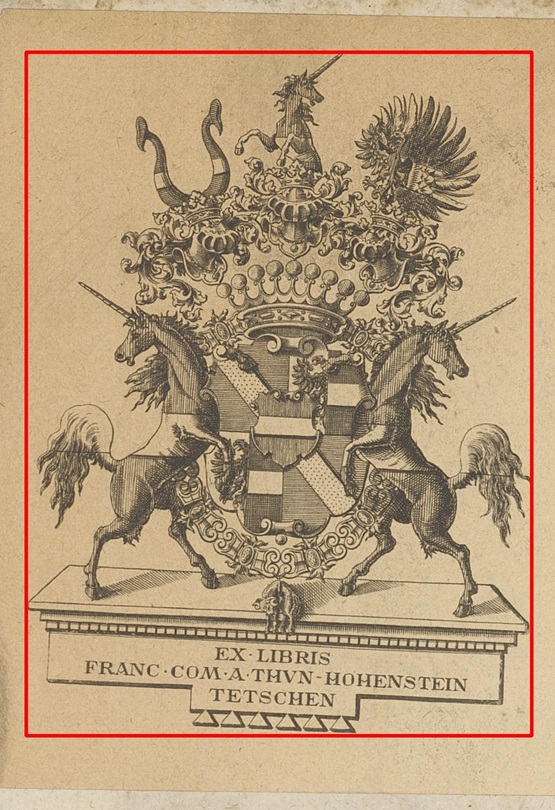}
        \caption{Exlibris}
        \label{fig:category_exlibris}
    \end{subfigure}
    \hfill
    \begin{subfigure}{0.515\linewidth}
        \centering
        \includegraphics[width=\linewidth]{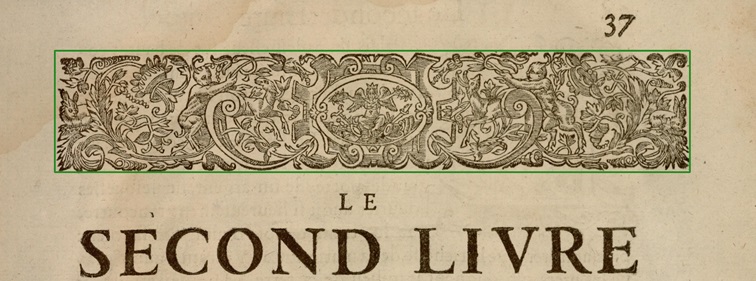}
        \includegraphics[width=\linewidth]{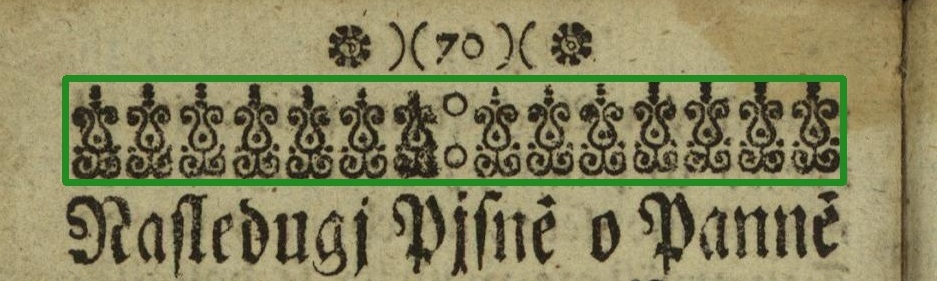}
        \caption{Frieze}
        \label{fig:category_frieze}
    \end{subfigure}
    \\ \vspace{0.2cm}
    \begin{subfigure}{0.475\linewidth}
        \centering
        \includegraphics[width=0.48\linewidth]{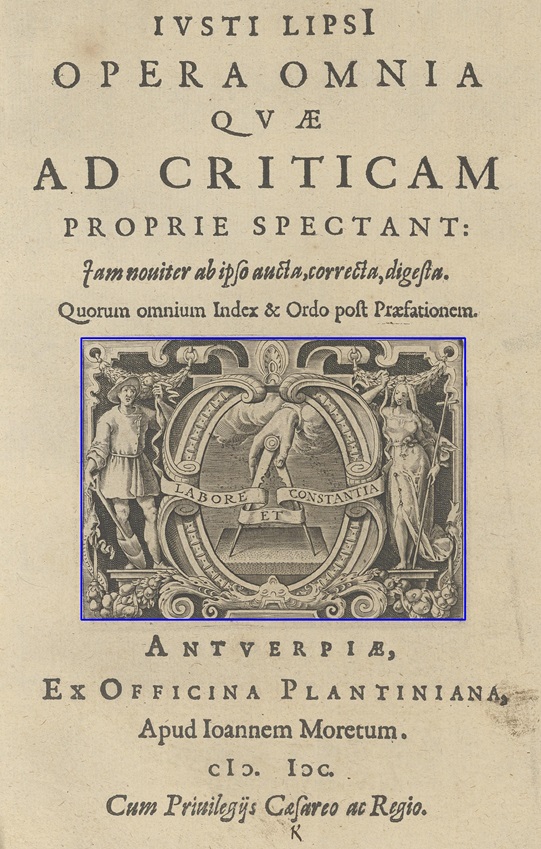}
        \includegraphics[width=0.48\linewidth]{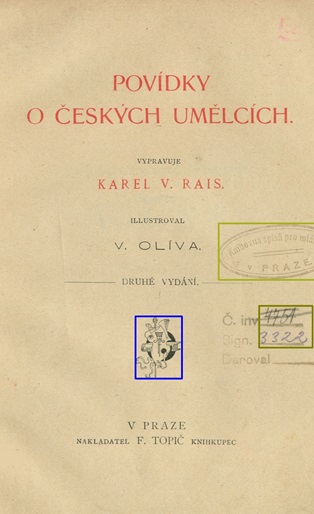}
        \caption{Signet}
        \label{fig:category_signet}
    \end{subfigure}
    \hfill
    \begin{subfigure}{0.51\linewidth}
        \centering
        \includegraphics[width=0.48\linewidth]{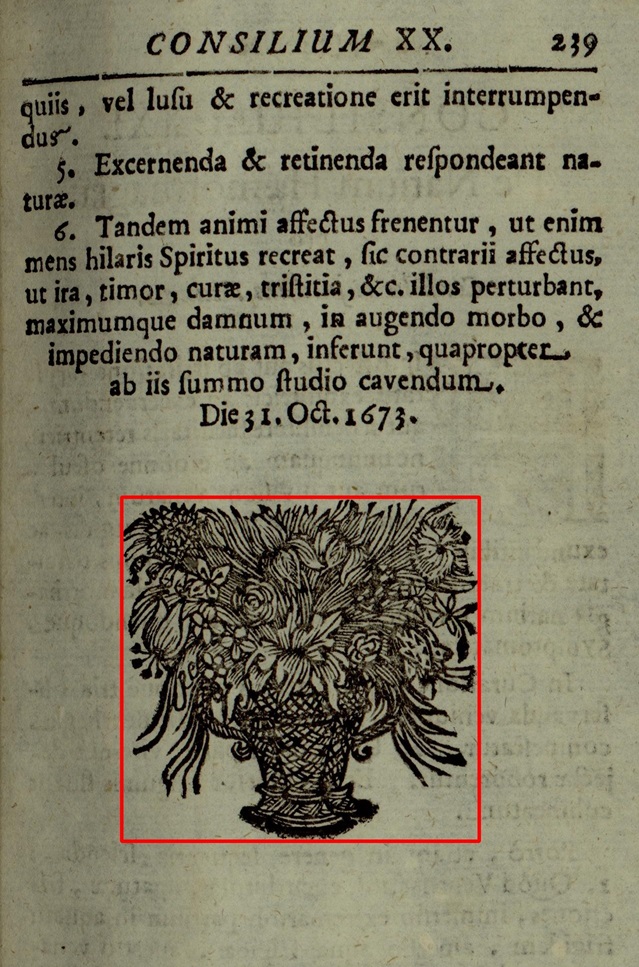}
        \includegraphics[width=0.48\linewidth]{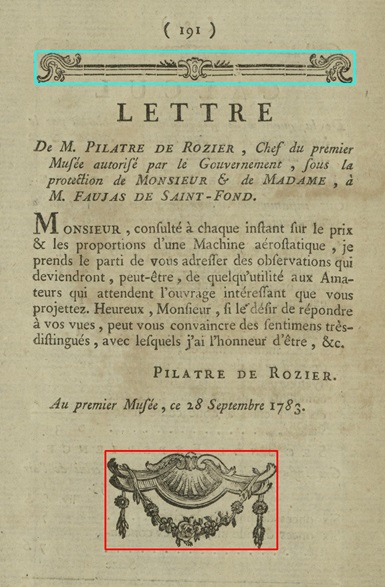}
        \caption{Vignette}
        \label{fig:category_vignette}
    \end{subfigure}
    \caption{Examples of selected element categories, which are less common and mostly related to historical documents.
    These categories are highlighted by bounding-boxes in orange (Exlibris), green (Frieze), blue (Signet), and red (Vignette).}
    \label{fig:categories}
\end{figure}

In the dataset we follow the Czech Methodology of image document processing~\cite{jirousek_metodika_2024}, which defines 25 different categories of non-textual elements.
Examples of less common categories are shown in Figure~\ref{fig:categories}.
These categories are typically related to historical documents and they are mostly important for historians, librarians, and other researchers.
The categories and their short descriptions are:
\begin{enumerate}
    \item \textbf{Chemical formula}: elements containing the composition and structure of chemical compounds or illustrations of chemical reactions, including reactants, products, and potential conditions.
    \item \textbf{Symbol}: visual elements that represent an idea, entity, or value through widely recognized signs, including icons, pictograms, logos, and heraldic emblems. These elements have historical and cultural significance and appear in various contexts, from documents to visual identity systems.
    \item \textbf{Exlibris}: a book owner's mark, typically a printed or handwritten label placed inside a book, often featuring the owner's name along with artistic elements such as figures, symbols, plant motifs, or heraldic emblems.
    \item \textbf{Photograph}: photographic images that capture real-world scenes with visual fidelity, regardless of printing technique. Exceptions include hand-crafted reproductions (e.g., woodcuts) where significant stylization alters the original image.
    \item \textbf{Geometric drawing}: elements featuring shapes, such as points, lines, triangles, squares, and more complex figures like circles, ellipses, cubes, and spheres, illustrate both planar and spatial forms using usually simple, drawn lines and annotations.
    \item \textbf{Chart}: any visual representation illustrating quantitative relationships between related variables using a combination of points, lines, numbers, symbols, and colors, typically within a coordinate system. 
    \item \textbf{Initial}: prominent letters at the beginning of text segments, often distinguished by size, decoration, or colour.
    \item \textbf{Cartoon}: elements combining images and optionally text to convey stories, satire, or humor. 
    \item \textbf{Map}: an element containing a reduced, generalized representation of the Earth’s surface, celestial bodies, or imaginary worlds, typically created using cartographic methods and symbols.
    \item \textbf{Mathematical formula}: an element combining symbols forming well-struc-tured single- or multi-line formulas according to the rules of a given mathematical context appearing on separate lines.
    \item \textbf{Sheet music}: an element containing one or more staves (typically five-lined) with or without musical symbols, used for graphical representation of music.
    \item \textbf{Image}: a non-text visual representation of an object, scene, or concept created through various artistic or technical means. This category also serves as a fallback for visual elements that do not fit into any other predefined category.
    \item \textbf{Other decoration}: elements including ornamental typographic elements such as borders, frames, and decorative strips that enhance the aesthetic appeal of the text.
    \item \textbf{Other technical drawing}: an element containing drawings, outlines, sections, or cross-sections of objects, created according to technical drawing principles, that does not fit into other categories of technical drawings (Floor plans, Diagrams, and Geometric drawings).
    \item \textbf{Decorative inscription}: an element with text written or printed in a decorative or artistic style that cannot be easily processed using standard OCR tools.
    \item \textbf{Technical drawing}: a top-down projection of a building, typically illustrating its spatial arrangement and structural layout.
    \item \textbf{Barcode}: an element carrying coded information used for identification in the management of library collections. The primary significance of this category lies in preventing confusion with other categories of graphic elements.
    \item \textbf{Stamp}: provenance marks indicating ownership and institutional usage of a document. They can be textual (containing only text) or graphic (including visual elements).
    \item \textbf{Advertisement}: an element visually promoting products, services, events, or ideas and is distinctly framed as a separate graphic unit.
    \item \textbf{Handwritten inscription}: handwritten annotations that are not part of the main content of the document, including marginalia, glosses, provenance notes, and personal records such as family chronicles.
    \item \textbf{Schema}: a simplified graphical representation that illustrates components of a system or process and their relationships, typically used for visualizing and organizing abstract concepts.
    \item \textbf{Signet}: a graphic mark with both identification and decorative functions, used by printers, publishers, and booksellers. Typically found on the title page or as part of the explicit (text added at the end of a book), it often includes a printer's monogram and motifs like torches, anchors, or biblical and mythological figures, serving as a visual reference to the printer's identity.
    \item \textbf{Table}: an element with a structured presentation of information or data organized into rows and columns.
    \item \textbf{Vignette}: a type of ornamental, emblematic, or calligraphic book decoration that can take various shapes. It is most commonly found on the title page between the title and the publisher's information or at the end of a chapter.
    \item \textbf{Frieze}: an elongated book decoration with aesthetic, optical, and sometimes symbolic functions, typically found at the top of pages. It features a central ornamental motif with symmetrical mirror-image sides.
\end{enumerate}

\subsection{Ground-truth creation}
We used Label Studio to annotate the collected pages, structuring the entire annotation workflow into four iterations.
In the first iteration, annotators manually created annotations for a subset of pages (4k pages).
In the second and third iterations, annotators worked with additional subsets of pages (2k and 1,5k pages, respectively), but this time assisted by predictions from a YOLO detector trained on previously annotated subsets.
These automatically generated proposals accelerated annotation, as annotators only needed to correct existing predictions rather than annotate from scratch.
By the end of the third iteration, annotations were complete for all collected pages.

In the final iteration, we verified the dataset's consistency.
We divided the dataset into two halves, trained a YOLO detector separately on each half, and generated predictions for their complementary halves.
We then identified pages where predictions significantly deviated from the ground truth, reviewing and correcting annotations as necessary.
Differences between predictions and annotations were evaluated based on geometric discrepancies, incorrect classifications, and prediction confidence scores.

\subsection{Final dataset}

\begin{figure}[t]
    \centering
    \begin{subfigure}{0.48\linewidth}
        \centering
        \includegraphics[width=\linewidth]{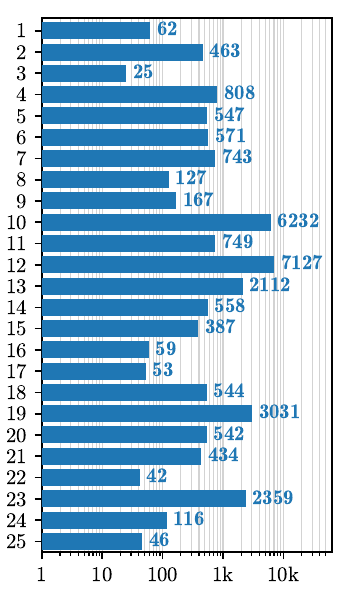}
        \caption{Dataset distribution}
        \label{fig:distribution_all}
    \end{subfigure}
    \hfill
    \begin{subfigure}{0.48\linewidth}
        \centering
        \includegraphics[width=\linewidth]{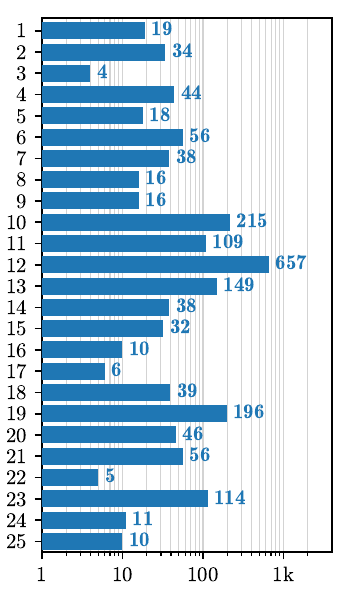}
        \caption{Test subset distribution}
        \label{fig:distribution_test}
    \end{subfigure}
    \caption{Distribution of the number of annotated elements for each category in the entire dataset and in the test subset. The number on the left-hand side indicates the category number identical to the listing in the text above. Note the logarithmic scale of the x-axis.}
    \label{fig:distribution}
\end{figure}

The final dataset contains $27\,904$ annotated elements on $6\,726$ pages and  824 pages without any annotated element.
Each element is represented by a tightly fitted axis-aligned bounding box (AABB).
The distribution of the number of annotations for each category is shown in Figure~\ref{fig:distribution_all}.
The dataset is divided into development and test subsets.
The further division of the development subset into the traditional training and validation subsets is left to the users.
The test subset contains 600 pages (21 pages without any annotated element) coming from the Czech subset.
We selected the pages of the test subset so that the relative frequency of each category was as similar to the entire set as possible.
The resulting test set contains a total of $1\,938$ annotated elements.
The distribution of the test subset is shown in Figure~\ref{fig:distribution_test}.

The dataset is publicly available on Zenodo platform\footnote{\url{https://doi.org/10.5281/zenodo.12788419}}.
It consists of pages from the Czech Digital Library, a text file containing a list of pages from the other datasets, and ground-truth text files in YOLO format for all pages.
Since the test subset contains only pages from the Czech subset, it is possible to use the dataset even without the need to obtain pages from the extra datasets.

\subsection{Limits of the dataset}
\label{sec:limits}

\begin{figure}[t]
    \centering
    \begin{subfigure}{0.48\linewidth}
        \centering
        \includegraphics[height=5cm]{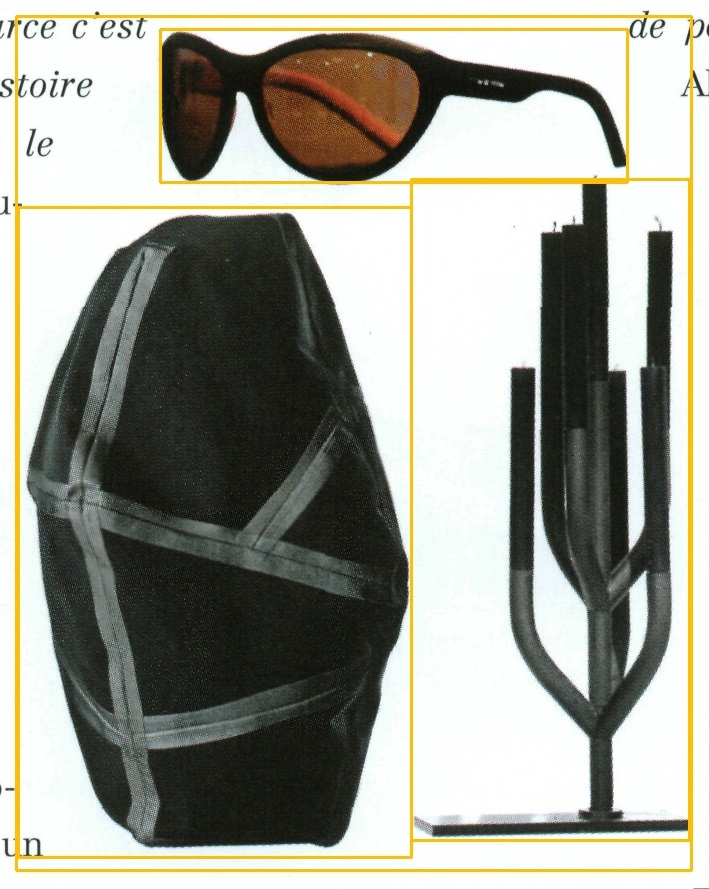}
        \caption{(Multiple) Images/photographs}
        \label{fig:limits_images_photographs}
    \end{subfigure}
    \hfill
    \begin{subfigure}{0.48\linewidth}
        \centering
        \includegraphics[height=5cm]{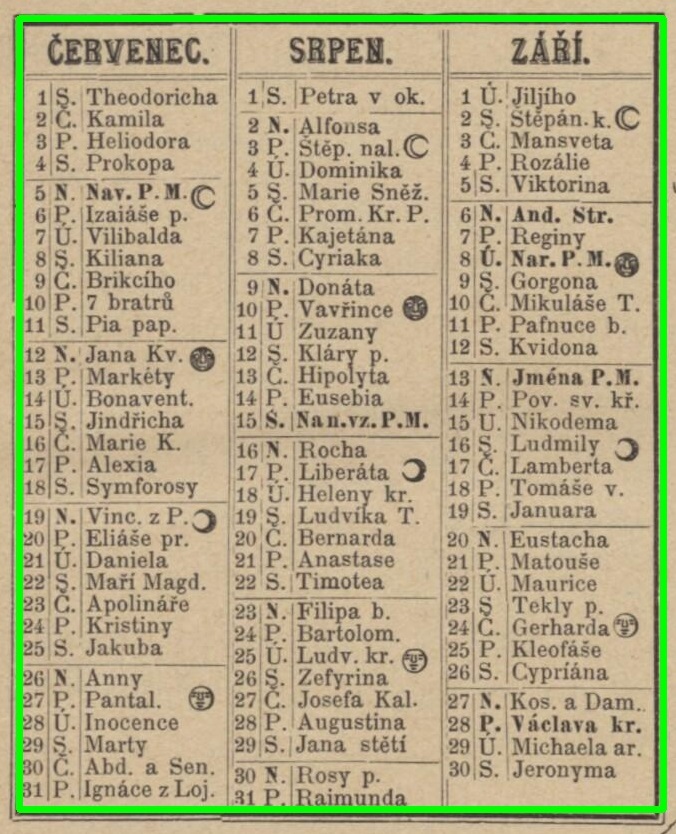}
        \caption{Tables}
        \label{fig:limits_tables}
    \end{subfigure}
    \\ \vspace{0.2cm}
    \begin{subfigure}{1.0\linewidth}
        \centering
        \includegraphics[width=0.42\linewidth]{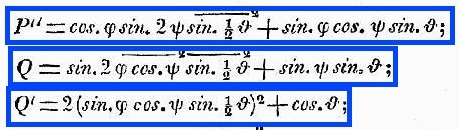}
        \hspace{0.5cm}
        \includegraphics[width=0.48\linewidth]{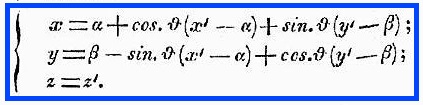}
        
        \caption{Equations}
        \label{fig:limits_equations}
    \end{subfigure}
    
    \caption{Examples of dataset limitations and ambiguities. For some elements, the class might be ambiguous (Figure~\ref{fig:limits_images_photographs}), or it might be unclear how it should be annotated -- whether enclose the entire element into a single annotation or use multiple instances (Figures~\ref{fig:limits_tables} and \ref{fig:limits_equations}).}
    \label{fig:limits}
\end{figure}

The AnnoPage Dataset has certain limitations that naturally arise with complex documents.
One primary limitation is category ambiguity, particularly when distinguishing between closely related classes. 
For instance, differentiating between images and photographs or various types of drawings can be problematic, as the boundaries between these categories are often unclear. 
A notable example is a product image that was initially photographed and later cropped -- such a case could be classified as either a photograph or an image, depending on the context and interpretation (see Figure~\ref{fig:limits_images_photographs}).

Another challenge arises in the annotation of decorative elements, especially with surrounding text. 
In some cases, decorative frames can be annotated as a single element, even if it contains text, while in other scenarios they can be split into separate elements, as it often depends on interpretation and context.
Similar ambiguity applies to tables, where multiple tables positioned closely together might be perceived as a single structure or as separate entities (see Figure~\ref{fig:limits_tables}). 
Additionally, text formatted in a tabular style -- lacking visible cell borders -- may resemble a table but not function as one, making classification difficult.
Mathematical and chemical formulas can also cause ambiguity, as it might be unclear whether two parts of a formula should be annotated as a single or multiple elements (see Figure~\ref{fig:limits_equations}).
Similarly, hierarchical structures within images are also challenging, as an image may contain sub-images that could be considered distinct elements on their own (see Figure~\ref{fig:limits_images_photographs}).

During the annotation process, these challenging cases were discussed and decided among the expert librarian annotators with respect to the Czech methodology of image document processing.

\subsection{Evaluation protocol}
The evaluation protocol of the AnnoPage Dataset follows the standard evaluation protocol for object detection.
The primary evaluation metrics are the mAP@50 and mAP@50-95.
The mAP (mean Average Precision) is calculated over the set of classes (the defined element categories) with a given IoU (intersection over union) threshold.
For mAP@50 the IoU threshold is set to 0.5 while the mAP@50-95 is calculated as an average of mAPs with IoU threshold ranging from 0.5 to 0.95 with step 0.05.

\section{Detection baselines}
\label{sec:baselines}
In this section, we present baseline methods and their results on the AnnoPage Dataset.
We trained YOLO~\cite{yolo11_ultralytics} and DETR~\cite{carion_end--end_2020} models on the development set and evaluated the best-performing models on the test set.
More specifically, we split the development set into training and validation in the same manner as in the case of the test set, i.e. we aimed to obtain a similar distribution of the labels in the validation set as in the entire development set.
As a result, we obtained a validation set of 509 pages with $1\,926$ annotated elements.
To assess the contribution of the Czech data, we also trained models only on the Czech subset and compared them with models trained on the full dataset.
First, we experimented with DETR and various YOLO models with fixed resolution of input images.
We also investigated the effect of different image resolutions on the detector performance.

We trained YOLO model variants \texttt{YOLO11n}, \texttt{YOLO11s}, \texttt{YOLO11m}, and \texttt{YOLO11l} using the Ultralytics package~\cite{yolo11_ultralytics}.
We took the publicly available pre-trained models and we optimized them for 250 epochs using Adam optimizer with base learning rate set to $2\times10^{-4}$.
For the DETR model, we used the implementation available on HuggingFace\footnote{\url{https://huggingface.co/docs/transformers/model_doc/detr}}.
More specifically, we took the pre-trained \texttt{facebook/detr-resnet-50}\footnote{\url{https://huggingface.co/facebook/detr-resnet-50}} model and we optimized it for 250 epochs with learning rate set to $5\times10^{-5}$
Here, we trained all the detectors with a fixed image resolution of $1\,024 \times 1\,024$ pixels.

The results of the trained models on the test set are summarized in Table~\ref{tab:baseline_results}.
The results show that \texttt{YOLO11m} and \texttt{YOLO11l} perform the best and their results are very similar on both the Czech subset and the full dataset.
When comparing the results between the Czech subset and the full dataset, it can be seen that the pages from the extra datasets bring improvements and help the models to generalize better by exposing them to a broader range of variability, although the Czech subset alone already provides  solid baseline performance.
However, the values of both metrics suggest that there is room for improvement.
On the other hand, the DETR performed the worst, probably due to the fact that Transformers generally require very large datasets to optimize well~\cite{oquab2023dinov2,radford2023robust,dubey2024llama}.

\begin{table}[t]
    \centering
    \caption{%
        Results of object detection baselines. 
        We report the mAP@50 and mAP@50-95 metrics on the test set  of models trained on the Czech subset and the full dataset.
    }\label{tab:baseline_results}
    \begin{tabular}{
        @{\extracolsep{4pt}}@{\kern\tabcolsep}
        lccccc}
        \toprule
        \multirow{2}{*}{Model} & \multicolumn{2}{c}{Czech subset} & & \multicolumn{2}{c}{Full dataset} \\ \cline{2-3} \cline{5-6} \addlinespace[0.1cm]
            & mAP@50 & mAP@50-95 & & mAP@50 & mAP@50-95 \\
        \midrule
        YOLO11n & 0.595 &  0.531 & &  0.616 &     0.558 \\ \addlinespace[0.1cm]
        YOLO11s & 0.567 & 0.513 & &  0.641 &     0.585 \\ \addlinespace[0.1cm]
        YOLO11m & 0.592 & 0.541 & &  \textbf{0.658} &     0.598 \\ \addlinespace[0.1cm]
        YOLO11l & \textbf{0.607} & \textbf{0.547} & &  0.657 &     \textbf{0.602} \\ \addlinespace[0.1cm]
        DETR     & 0.438 & 0.356 & &  0.458 &    0.377  \\
        \bottomrule
    \end{tabular}
\end{table}

\begin{figure}[]
    \centering
    \begin{subfigure}{\linewidth}
        \centering
        \includegraphics[height=7.5cm]{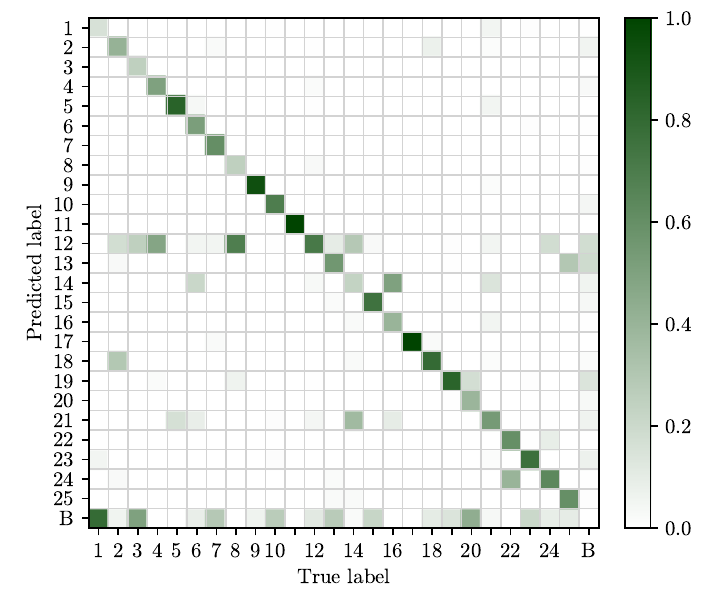}
        \caption{Czech subset}
        \label{fig:confusion_matrix_czech}
    \end{subfigure}
    \begin{subfigure}{\linewidth}
        \centering
        \includegraphics[height=7.5cm]{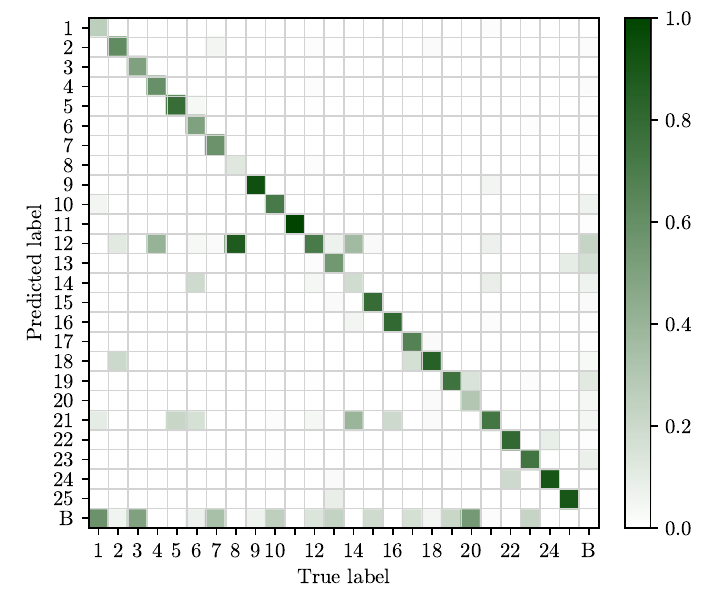}
        \caption{Full dataset}
        \label{fig:confusion_matrix_full}
    \end{subfigure}    
    \caption{Confusion matrices for \texttt{YOLO11l} models trained on the Czech subset (Fig.~\ref{fig:confusion_matrix_czech}) and the full dataset (Fig.~\ref{fig:confusion_matrix_full}). The numbers on the left and bottom indicate the categories defined in the dataset description (Section~\ref{sec:categories}). The last row and column (labeled 'B') is the category called 'Background' -- the bottom row represents false negatives and the last column represents false positives. Each matrix is column-wise normalized.}
    \label{fig:confusion_matrix}
\end{figure}

In Figure~\ref{fig:confusion_matrix}, we present a confusion matrices for the \texttt{YOLO11l} model trained on the Czech subset and the full dataset and evaluated on the test set.
The confusion matrix shows that the model most often confuses photographs (4) and cartoons (8) with images (12) and other technical drawings (14) with schemas (21).
However, it should be mentioned that especially the cartoon is relatively less frequent in the test set, so any confusion is then even more pronounced in the matrix.
The matrix also shows that the model quite often misses elements (B), especially chemical formulas (1), exlibris (3) and handwritten inscriptions (20).

\begin{figure}[!ht]
    \centering
    \includegraphics[height=5.5cm]{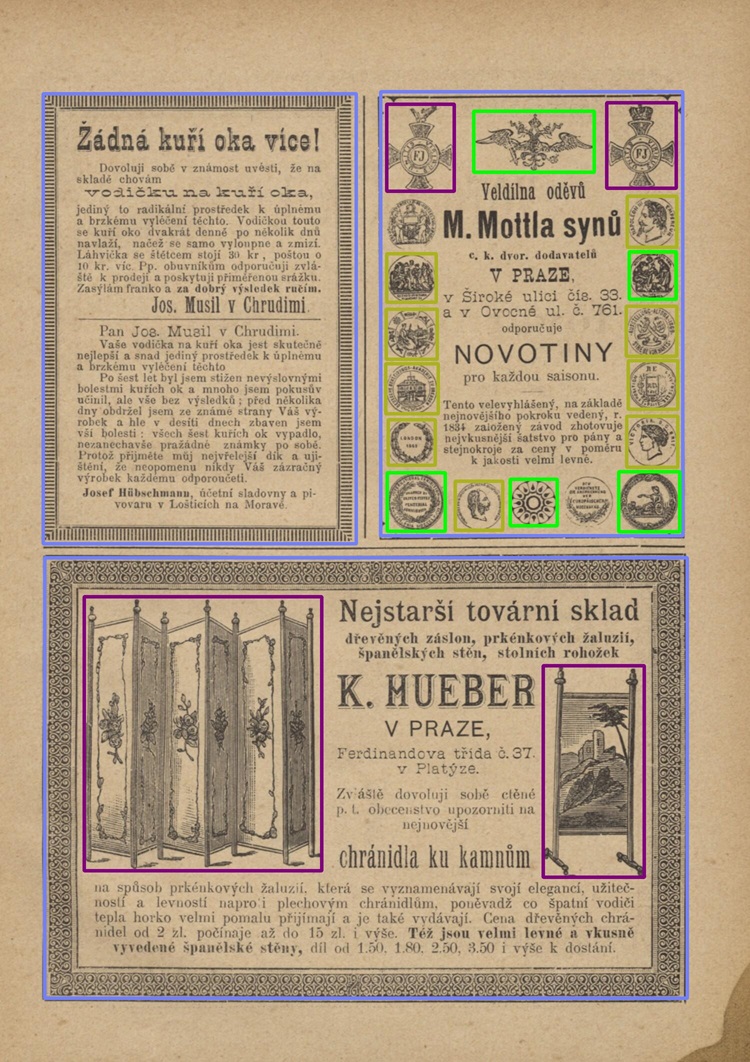}
    \hfill
    \includegraphics[height=5.5cm]{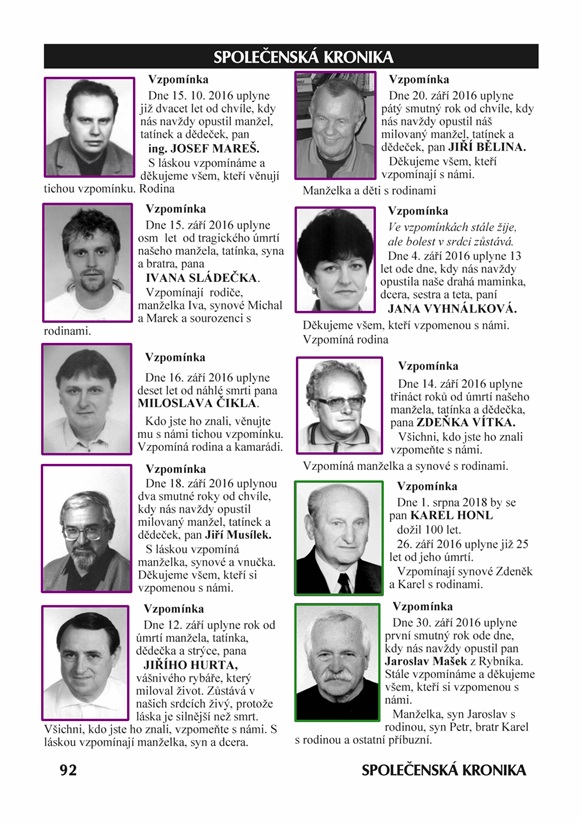}
    \hfill
    \includegraphics[height=5.5cm]{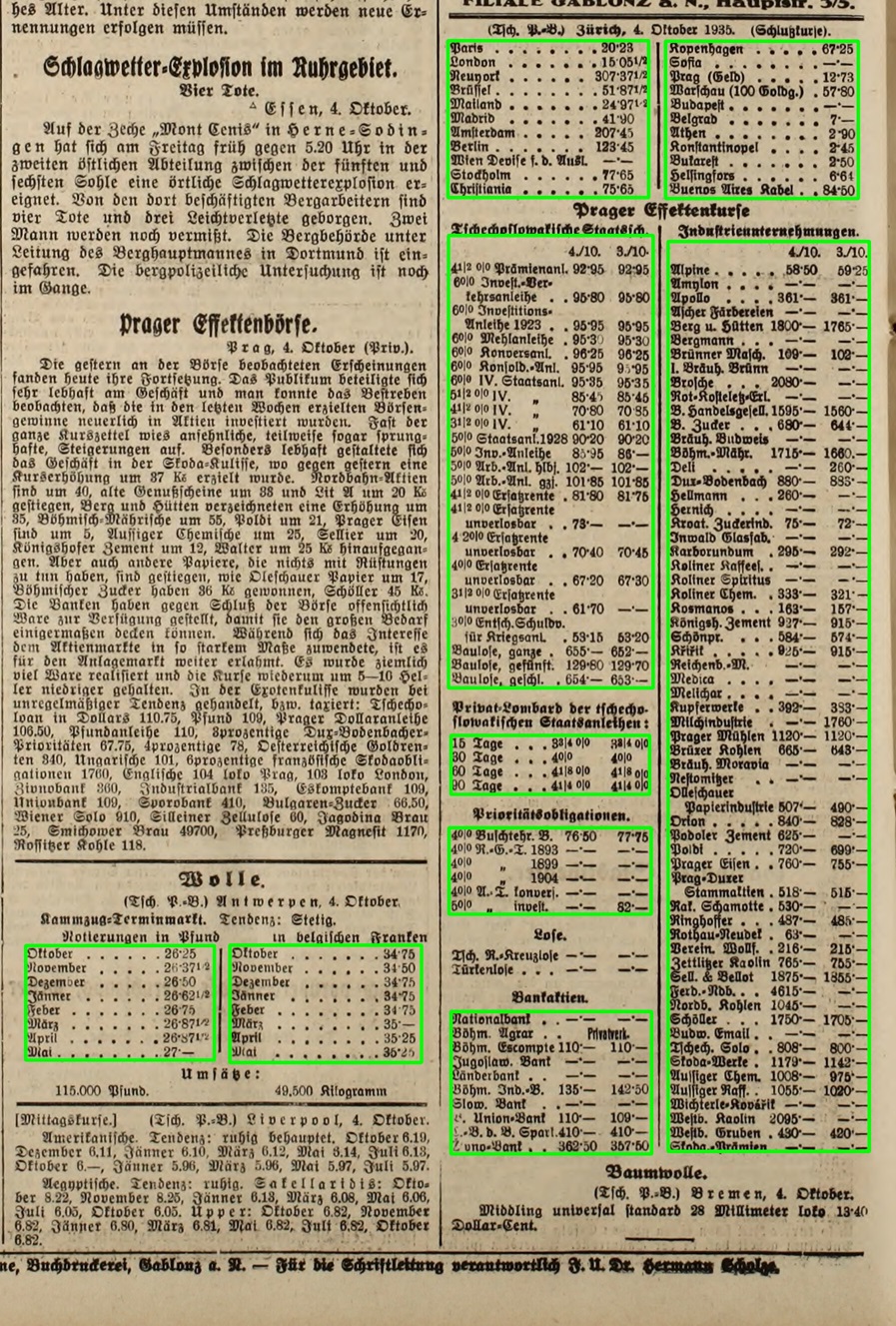}
    \\ \vspace{0.1cm}
    \includegraphics[height=5.5cm]{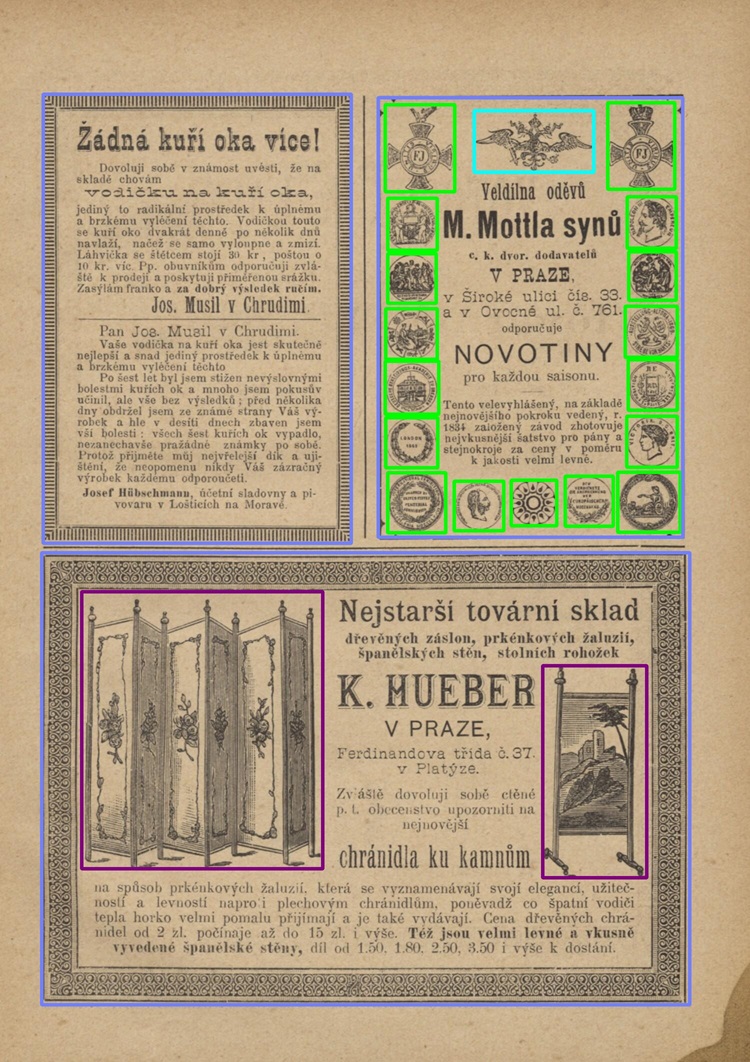}
    \hfill
    \includegraphics[height=5.5cm]{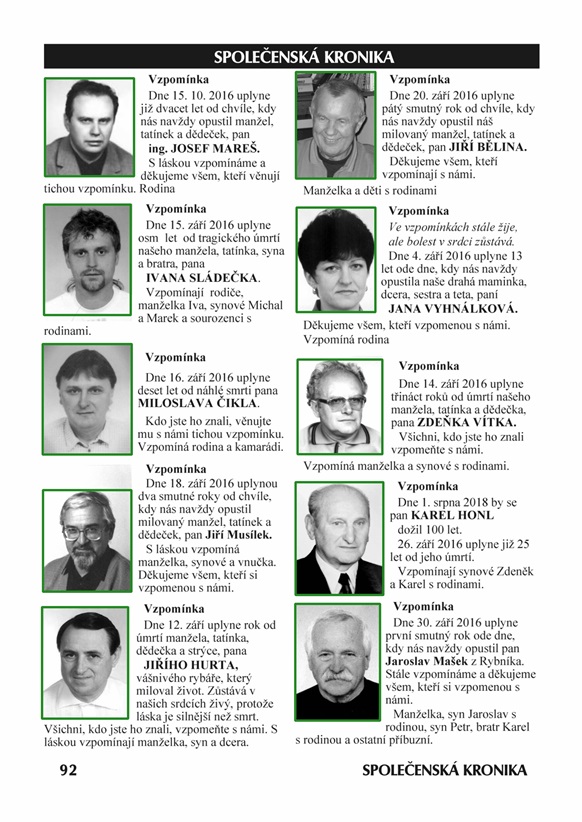}
    \hfill
    \includegraphics[height=5.5cm]{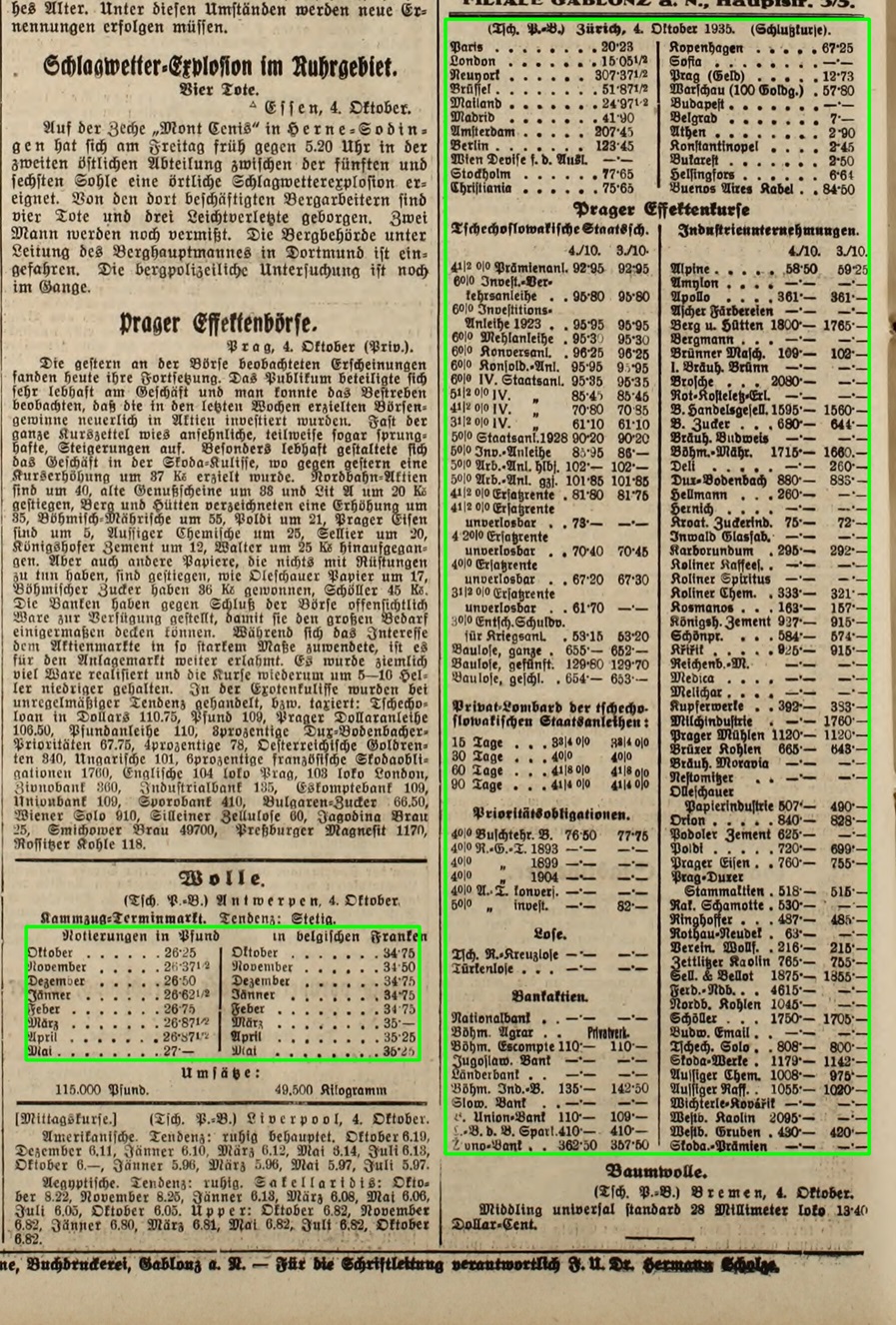}
    \caption{Example of erroneous predictions produced by \texttt{YOLO11l} model trained on the full dataset. The first row contains pages with rendered predictions and the second row contains rendered ground-truth.
    Examples on the left and in the middle contain objects with wrongly identified classes while the right example contains more fine grained predictions than ground-truth.}
    \label{fig:errors}
\end{figure}

Figure~\ref{fig:errors} contains example of pages with erroneous predictions of \texttt{YOLO11l} model trained on the full dataset.
The examples on the left and in the middle show objects for which the categories were wrongly predicted. 
The sample on the right contains correctly identified categories (tables), but the model produced more fine grained detections than ground-truth.
This is related to the limits of the dataset described in Section~\ref{sec:limits}, where it is not clear whether it is a single table or multiple smaller tables.

We also experimented with different resolutions of the input images.
We trained the \texttt{YOLO11m} model with the same experimental setup as in the previous experiments, but with a changed image size.
In addition to the resolution $1024\times1024$ pixels, we used the resolutions $800\times800$ pixels and $1280\times1280$ pixels.
The results are in Table~\ref{tab:resolution_results} and they show that the resolution of $1024 \times 1024$ pixels is the most suitable.

\begin{table}[t]
    \centering
    \caption{%
        Results of \texttt{YOLO11m} at different resolutions. We report the mAP@50 and mAP@50-95 metrics of each model on the test set.
    }\label{tab:resolution_results}
    \begin{tabular}{
        @{\extracolsep{4pt}}@{\kern\tabcolsep}
        rcc}
        \toprule
        Resolution & mAP@50 & mAP@50-95 \\
        \midrule
         800 px & 0.622 &     0.561 \\ \addlinespace[0.1cm]
        1024 px & \textbf{0.658} &     \textbf{0.598} \\ \addlinespace[0.1cm]
        1280 px & 0.651 &     0.590 \\
        \bottomrule
    \end{tabular}
\end{table}

\section{Conclusions}
In this paper, we introduced the AnnoPage Dataset, a novel collection of $7\,550$ mainly historical document pages annotated with axis-aligned bounding boxes for 25 categories of non-textual elements.
The annotations were created by expert librarians based on the Czech Methodology of image document processing.
We described the entire creation process, the defined categories, limits of the dataset, and its evaluation protocol.

To establish a baseline for future research, we evaluated the dataset using YOLO and DETR object detection models, demonstrating its applicability for document layout analysis and object detection tasks. 
The dataset is publicly available on Zenodo, along with annotations in the YOLO format and pre-defined test subset, encouraging further research and development in historical document processing.
We hope that the AnnoPage Dataset will serve as a valuable resource for further researcher in document image analysis, object detection, and related fields.

\begin{credits}
\subsubsection{\ackname}
This work has been supported by the Ministry of Culture Czech Republic in NAKI III project Orbis Pictus – book revival for cultural and creative sectors (DH23P03OVV033).

\subsubsection{\discintname}
The authors have no competing interests to declare that are
relevant to the content of this article.
\end{credits}

\bibliographystyle{splncs04}
\bibliography{references}

\begin{thebibliography}{10}
\providecommand{\url}[1]{\texttt{#1}}
\providecommand{\urlprefix}{URL }
\providecommand{\doi}[1]{https://doi.org/#1}

\bibitem{antonacopoulos_realistic_2009}
Antonacopoulos, A., Bridson, D., Papadopoulos, C., Pletschacher, S.: A realistic dataset for performance evaluation of document layout analysis. In: 2009 10th International Conference on Document Analysis and Recognition. pp. 296--300 (2009), {ISSN}: 2379-2140

\bibitem{auer_icdar_2023}
Auer, C., Nassar, A., Lysak, M., Dolfi, M., Livathinos, N., Staar, P.: {ICDAR} 2023 competition on robust layout segmentation in corporate documents. In: Fink, G.A., Jain, R., Kise, K., Zanibbi, R. (eds.) Document Analysis and Recognition - {ICDAR} 2023. Springer Nature Switzerland (2023)

\bibitem{boillet_horae_2019}
Boillet, M., Bonhomme, M.L., Stutzmann, D., Kermorvant, C.: {HORAE}: an annotated dataset of books of hours. In: Proceedings of the 5th International Workshop on Historical Document Imaging and Processing. pp. 7--12. {HIP} '19, Association for Computing Machinery (2019)

\bibitem{buttner_cordeep_2022}
Büttner, J., Martinetz, J., El-Hajj, H., Valleriani, M.: {CorDeep} and the sacrobosco dataset: Detection of visual elements in historical documents. Journal of Imaging  \textbf{8}(10), ~285 (2022), number: 10 Publisher: Multidisciplinary Digital Publishing Institute

\bibitem{carion_end--end_2020}
Carion, N., Massa, F., Synnaeve, G., Usunier, N., Kirillov, A., Zagoruyko, S.: End-to-end object detection with transformers. In: Vedaldi, A., Bischof, H., Brox, T., Frahm, J.M. (eds.) Computer Vision – {ECCV} 2020. pp. 213--229. Springer International Publishing (2020)

\bibitem{cheng_m6doc_2023}
Cheng, H., Zhang, P., Wu, S., Zhang, J., Zhu, Q., Xie, Z., Li, J., Ding, K., Jin, L.: M6doc: A large-scale multi-format, multi-type, multi-layout, multi-language, multi-annotation category dataset for modern document layout analysis. pp. 15138--15147

\bibitem{clausner_icdar2019_2019}
Clausner, C., Antonacopoulos, A., Pletschacher, S.: {ICDAR}2019 competition on recognition of documents with complex layouts - {RDCL}2019. In: 2019 International Conference on Document Analysis and Recognition ({ICDAR}). pp. 1521--1526 (2019), {ISSN}: 2379-2140

\bibitem{clausner_enp_2015}
Clausner, C., Papadopoulos, C., Pletschacher, S., Antonacopoulos, A.: The {ENP} image and ground truth dataset of historical newspapers. In: 2015 13th International Conference on Document Analysis and Recognition ({ICDAR}). pp. 931--935 (2015)

\bibitem{diem_cbad_2019}
Diem, M., Kleber, F., Sablatnig, R., Gatos, B.: {cBAD}: {ICDAR}2019 competition on baseline detection. In: 2019 International Conference on Document Analysis and Recognition ({ICDAR}). pp. 1494--1498 (2019), {ISSN}: 2379-2140

\bibitem{dubey2024llama}
Dubey, A., Jauhri, A., Pandey, A., Kadian, A., Al-Dahle, A., Letman, A., Mathur, A., Schelten, A., Yang, A., Fan, A., et~al.: The llama 3 herd of models. arXiv preprint arXiv:2407.21783  (2024)

\bibitem{gao_icdar_2019}
Gao, L., Huang, Y., Déjean, H., Meunier, J.L., Yan, Q., Fang, Y., Kleber, F., Lang, E.: {ICDAR} 2019 competition on table detection and recognition ({cTDaR}). In: 2019 International Conference on Document Analysis and Recognition ({ICDAR}). pp. 1510--1515 (2019), {ISSN}: 2379-2140

\bibitem{jaume2019funsd}
Jaume, G., Ekenel, H.K., Thiran, J.P.: Funsd: A dataset for form understanding in noisy scanned documents. In: 2019 International Conference on Document Analysis and Recognition Workshops (ICDARW). vol.~2, pp.~1--6. IEEE (2019)

\bibitem{jirousek_metodika_2024}
Jiroušek, V., Pavčík, F., Hrzinová, J., Kersch, F., Jebavý, F., Hradiš, M., Bežová, M., Lehečka, B., Žabička, P., Kišš, M., Škvrňák, J., Bednařík, P., Fremrová, K., Dvořáková, M., Smolka, L., Lhoták, M.: Metodika zpracování obrazových dokumentů, \url{http://www.nusl.cz/ntk/nusl-668880}, publisher: Knihovna {AV} ČR, Národní 3, 115 22 Praha 1, http://www.lib.cas.cz/

\bibitem{yolo11_ultralytics}
Jocher, G., Qiu, J.: Ultralytics yolo11 (2024), \url{https://github.com/ultralytics/ultralytics}

\bibitem{kodym_page_2021}
Kodym, O., Hradiš, M.: Page layout analysis system for unconstrained historic documents. In: Lladós, J., Lopresti, D., Uchida, S. (eds.) Document Analysis and Recognition – {ICDAR} 2021. pp. 492--506. Lecture Notes in Computer Science, Springer International Publishing (2021)

\bibitem{li_mask_2023}
Li, F., Zhang, H., Xu, H., Liu, S., Zhang, L., Ni, L.M., Shum, H.Y.: Mask {DINO}: Towards a unified transformer-based framework for object detection and segmentation. In: 2023 {IEEE}/{CVF} Conference on Computer Vision and Pattern Recognition ({CVPR}). pp. 3041--3050 (2023), {ISSN}: 2575-7075

\bibitem{li_tablebank_2020}
Li, M., Cui, L., Huang, S., Wei, F., Zhou, M., Li, Z.: {TableBank}: Table benchmark for image-based table detection and recognition. In: Calzolari, N., Béchet, F., Blache, P., Choukri, K., Cieri, C., Declerck, T., Goggi, S., Isahara, H., Maegaard, B., Mariani, J., Mazo, H., Moreno, A., Odijk, J., Piperidis, S. (eds.) Proceedings of the Twelfth Language Resources and Evaluation Conference. pp. 1918--1925. European Language Resources Association (2020)

\bibitem{li_docbank_2020}
Li, M., Xu, Y., Cui, L., Huang, S., Wei, F., Li, Z., Zhou, M.: {DocBank}: A benchmark dataset for document layout analysis. In: Scott, D., Bel, N., Zong, C. (eds.) Proceedings of the 28th International Conference on Computational Linguistics. pp. 949--960. International Committee on Computational Linguistics (2020)

\bibitem{ma_hrdoc_2023}
Ma, J., Du, J., Hu, P., Zhang, Z., Zhang, J., Zhu, H., Liu, C.: {HRDoc}: dataset and baseline method toward hierarchical reconstruction of document structures. In: Proceedings of the Thirty-Seventh {AAAI} Conference on Artificial Intelligence and Thirty-Fifth Conference on Innovative Applications of Artificial Intelligence and Thirteenth Symposium on Educational Advances in Artificial Intelligence. {AAAI}'23/{IAAI}'23/{EAAI}'23, vol.~37, pp. 1870--1877. {AAAI} Press (2023)

\bibitem{mehri_hba_2017}
Mehri, M., Héroux, P., Mullot, R., Moreux, J.P., Coüasnon, B., Barrett, B.: {HBA} 1.0: A pixel-based annotated dataset for historical book analysis. In: Proceedings of the 4th International Workshop on Historical Document Imaging and Processing. pp. 107--112. {HIP} '17, Association for Computing Machinery (2017)

\bibitem{monnier_docextractor_2020}
Monnier, T., Aubry, M.: {docExtractor}: An off-the-shelf historical document element extraction. In: 2020 17th International Conference on Frontiers in Handwriting Recognition ({ICFHR}). pp. 91--96 (2020)

\bibitem{oquab2023dinov2}
Oquab, M., Darcet, T., Moutakanni, T., Vo, H., Szafraniec, M., Khalidov, V., Fernandez, P., Haziza, D., Massa, F., El-Nouby, A., et~al.: Dinov2: Learning robust visual features without supervision. arXiv preprint arXiv:2304.07193  (2023)

\bibitem{pfitzmann_doclaynet_2022}
Pfitzmann, B., Auer, C., Dolfi, M., Nassar, A.S., Staar, P.: {DocLayNet}: A large human-annotated dataset for document-layout segmentation. In: Proceedings of the 28th {ACM} {SIGKDD} Conference on Knowledge Discovery and Data Mining. pp. 3743--3751. {KDD} '22, Association for Computing Machinery (2022)

\bibitem{radford2023robust}
Radford, A., Kim, J.W., Xu, T., Brockman, G., McLeavey, C., Sutskever, I.: Robust speech recognition via large-scale weak supervision. In: International conference on machine learning. pp. 28492--28518. PMLR (2023)

\bibitem{shahab_open_2010}
Shahab, A., Shafait, F., Kieninger, T., Dengel, A.: An open approach towards the benchmarking of table structure recognition systems. In: Proceedings of the 9th {IAPR} International Workshop on Document Analysis Systems. pp. 113--120. {DAS} '10, Association for Computing Machinery (2010)

\bibitem{simistira_diva-hisdb_2016}
Simistira, F., Seuret, M., Eichenberger, N., Garz, A., Liwicki, M., Ingold, R.: {DIVA}-{HisDB}: A precisely annotated large dataset of challenging medieval manuscripts. In: 2016 15th International Conference on Frontiers in Handwriting Recognition ({ICFHR}). pp. 471--476. {ISSN}: 2167-6445

\bibitem{smock_pubtables-1m_2022}
Smock, B., Pesala, R., Abraham, R.: {PubTables}-1m: Towards comprehensive table extraction from unstructured documents. In: 2022 {IEEE}/{CVF} Conference on Computer Vision and Pattern Recognition ({CVPR}). pp. 4624--4632. {IEEE} (2022)

\bibitem{soboroff2022complex}
Soboroff, I.: Complex document information processing (cdip) dataset. National Institute of Standards and Technology  (2022)

\bibitem{tom_icdar_2023}
Tom, G., Mathew, M., Garcia-Bordils, S., Karatzas, D., Jawahar, C.: {ICDAR} 2023 competition on {RoadText} video text detection, tracking and recognition. In: Fink, G.A., Jain, R., Kise, K., Zanibbi, R. (eds.) Document Analysis and Recognition - {ICDAR} 2023. pp. 577--586. Springer Nature Switzerland (2023)

\bibitem{tschirschwitz_dataset_2022}
Tschirschwitz, D., Klemstein, F., Stein, B., Rodehorst, V.: A dataset for analysing complex document layouts in the digital humanities and its evaluation with krippendorff’s alpha. In: Andres, B., Bernard, F., Cremers, D., Frintrop, S., Goldlücke, B., Ihrke, I. (eds.) Pattern Recognition. pp. 354--374. Springer International Publishing (2022)

\bibitem{zhang_dino_2022}
Zhang, H., Li, F., Liu, S., Zhang, L., Su, H., Zhu, J., Ni, L.M., Shum, H.Y.: {DINO}: {DETR} with improved {DeNoising} anchor boxes for end-to-end object detection (2022)

\bibitem{zhao_doclayout-yolo_2024}
Zhao, Z., Kang, H., Wang, B., He, C.: {DocLayout}-{YOLO}: Enhancing document layout analysis through diverse synthetic data and global-to-local adaptive perception (2024)

\bibitem{zhong_publaynet_2019}
Zhong, X., Tang, J., Jimeno~Yepes, A.: {PubLayNet}: Largest dataset ever for document layout analysis. In: 2019 International Conference on Document Analysis and Recognition ({ICDAR}). pp. 1015--1022 (2019), {ISSN}: 2379-2140

\bibitem{zhu_deformable_2021}
Zhu, X., Su, W., Lu, L., Li, B., Wang, X., Dai, J.: Deformable {DETR}: Deformable transformers for end-to-end object detection (2021)

\bibitem{zottin_u-diads-bib_2024}
Zottin, S., De~Nardin, A., Colombi, E., Piciarelli, C., Pavan, F., Foresti, G.L.: U-{DIADS}-bib: a full and few-shot pixel-precise dataset for document layout analysis of ancient manuscripts. Neural Comput. Appl.  \textbf{36}(20),  11777--11789 (2024)

\end{thebibliography}

\end{document}